\documentclass{amia}
\begin{document}

% \title{Prompt Element Risk Factors for Hallucinations and Omissions in Mental-Health LLM Responses: A Matched Observational Study Using a Modular Prompt Generation System}

\title{Disentangling Prompt Element Level Risk Factors for Hallucinations and Omissions in Mental Health LLM Responses}

\title{Disentangling Prompt Element Level Risk Factors for Hallucinations and Omissions in Mental Health LLM Responses}

\author{
Congning Ni, PhD$^{1}$,
Sarvech Qadir, MS$^{2}$,
Bryan Steitz, PhD$^{1}$,
Mihir Sachin Vaidya, MS$^{1}$,
Qingyuan Song, MS$^{2}$,
Lantian Xia, BS$^{2}$,
Shelagh Mulvaney, PhD$^{2}$,
Siru Liu, PhD$^{1}$,
Hyeyoung Ryu, PhD$^{1}$,
Leah Hecht, PhD$^{3}$,
Amy Bucher, PhD$^{3}$,
Christopher Symons, PhD$^{3}$,
Laurie Novak, PhD$^{1}$,
Susannah L. Rose, PhD$^{1}$,
Murat Kantarcioglu, PhD$^{4}$,
Bradley Malin, PhD$^{1,2}$,
Zhijun Yin, PhD$^{1,2}$
}

\institutes{
$^{1}$Vanderbilt University Medical Center, Nashville, TN, USA;
$^{2}$Vanderbilt University, Nashville, TN, USA;
$^{3}$Lirio, Knoxville, TN, USA;
$^{4}$Virginia Tech, Blacksburg, VA, USA
}

\maketitle

\section*{Abstract}
\textit{Mental health concerns are often expressed outside clinical settings, including in high-distress help seeking, where safety-critical guidance may be needed. Consumer health informatics systems increasingly incorporate large language models (LLMs) for mental health question answering, yet many evaluations underrepresent narrative, high-distress inquiries. We introduce $UTCO$ (User, Topic, Context, Tone), a prompt construction framework that represents an inquiry as four controllable elements for systematic stress testing. Using 2,075 $UTCO$-generated prompts, we evaluated Llama 3.3 and annotated hallucinations (fabricated or incorrect clinical content) and omissions (missing clinically necessary or safety-critical guidance). Hallucinations occurred in 6.5\% of responses and omissions in 13.2\%, with omissions concentrated in crisis and suicidal ideation prompts. Across regression, element-specific matching, and similarity-matched comparisons, failures were most consistently associated with context and tone, while user-background indicators showed no systematic differences after balancing. These findings support evaluating omissions as a primary safety outcome and moving beyond static benchmark question sets.}

\section*{Introduction}

Major depression and generalized anxiety disorders remain among the leading causes of disability globally\cite{friedrich2017depression} and continue to place immense pressure on mental health services in many regions\cite{wang2007use,mahmud2021global}. Due to persistent barriers to accessing healthcare, including workforce shortages, high costs, and stigma\cite{endalamaw2024barriers,alam2022addressing}, people increasingly rely on digital tools to interpret symptoms and seek medical assistance\cite{bocking2022m}. As patients and caregivers increasingly seek mental health support through online resources and interactive systems outside clinical settings, these tools have become a component component of consumer health informatics. The World Health Organization has emphasized the scale of unmet needs and the urgency of strengthening mental health systems, particularly ensuring that people have timely access to appropriate support\cite{world2022world}.
Meanwhile, conversational large language models (LLMs) are increasingly being used to address mental health concerns\cite{elyoseph2024comparing}, due to their ability to provide instant responses and engage in conversational dialogue\cite{zhang2024s}. 
While recent research highlights the growing application of these models in areas such as anxiety, depression, relationship stress, and crisis situations\cite{omar2025exploring,bucur2024leveraging}, they also emphasize that current safeguards and evaluation standards lag behind the actual use of these models\cite{reddy2023evaluating,freyer2024future}. This gap is particularly of concern in the mental health field, as users may interpret the models' responses as clinical guidance\cite{yu2024don}.

In consumer-facing mental health information seeking, including use by members of the general public and by caregivers, family members, and peers seeking guidance for someone else, two types of LLM response failures are particularly important. First, \emph{hallucinations} are incorrect or fabricated statements that could mislead decisions about treatment, medication, or safety\cite{zhang2025llm}. Second, \emph{omissions} are failures to provide clinically important information or safety guidance that would reasonably be expected in response to a user's inquiry, including when professional help should be recommended\cite{ohde2025burden}. Prior research has documented hallucinations in medical large language model outputs and the impact of seemingly plausible, but incorrect, text on patient safety\cite{tung2023potential,tan2024chatgpt,li2025ethical}. However, many evaluation methods still rely on fixed benchmark question sets that do not systematically reflect key real-world nuances, including user-background context, variation in inquiry content and phrasing, and the need to integrate narrative context with safety considerations\cite{freyer2024future,reddy2023evaluating,chen2025evaluating}.

To address this limitation, we designed and implemented the $UTCO$ (User, Topic, Context, Tone) framework to conduct a structured, element-based evaluation of a state-of-the-art open-weight conversational LLM (Llama 3.3) for mental health counseling. This framework modularizes each prompt into four components: into \emph{$\mathbf{U}$ser}, \emph{$\mathbf{T}$opic}, \emph{$\mathbf{C}$ontext}, and \emph{T$\mathbf{o}$ne}. This design supports controlled stress testing across clinically relevant domains and communication styles while maintaining authenticity through naturalistic contextual elements. We selected Llama 3.3 because it is widely deployed as an open-weight model and can be evaluated reproducibly, enabling transparent inspection of failure modes without reliance on proprietary APIs. 
Although our primary evaluation focuses on a single model, prior work suggests that prompt vulnerabilities and failure-inducing prompt patterns can transfer across different LLM families\cite{landon2025variation}, motivating mechanism-focused analyses that can be tested in other systems.
Using 2,075 queries, we quantified instances of hallucination and omission through human annotations, and then applied a three-stage analytical strategy to investigate the following research questions: % aligned with common informatics objectives: (1) identifying prompt-level risk factors associated with failure; (2) testing whether demographic indicators were associated with failure after balancing other prompt content; and (3) describing the linguistic mechanisms in nearly identical prompts that yielded different results.

\begin{itemize}
    \item \textbf{RQ1}: Which prompt elements across $U$, $T$, $C$, and $O$ are associated with hallucinations and omissions?

    \item \textbf{RQ2}: When holding three $UTCO$ elements constant, how does varying the remaining element change the risk of hallucinations and omissions?

    \item \textbf{RQ3}: For near-identical $UTCO$ prompts with different outcomes, what linguistic patterns distinguish failure cases from non-failure controls?
\end{itemize}

In this study, we evaluate LLM behavior in consumer-facing mental health information seeking using the $UTCO$ framework, which enables systematic variation of user background, topic, context, and tone while controlling other prompt elements. We quantify two safety-relevant failure modes, hallucinations and omissions, and use complementary analyses to characterize prompt conditions associated with failure. By focusing on narrative, high-distress help-seeking language, this work aims to better align mental health LLM evaluation with the types of inquiries encountered in real-world consumer settings.

\section*{Related Work}

\textbf{Hallucinations and Omissions in Large Language Models. }
% LLMs have demonstrated unprecedented fluency and performance across language-generation tasks. However, their widespread deployment has also revealed fundamental reliability challenges, particularly in domains demanding high factual fidelity and completeness. 
% A central concern in applying LLMs in healthcare is hallucination, the generation of syntactically plausible text that is factually incorrect, unsupported, or fabricated. 
Prior surveys characterize hallucinations using taxonomies of error types %(e.g., factual errors, context collisions, temporal inconsistency, ethical violations) 
and related these failures to data, training, and inference factors \cite{alansari2025large, huang2025survey}. Hallucination rates and forms vary by model architecture, scale, pretraining corpora, and application context, indicating sensitivity to both model properties and task framing \cite{alansari2025large}. Clinical and safety-focused work also highlights omission as a distinct and consequential failure mode\cite{asgari2025framework}. %, including missing crucial information, failing to escalate risk, and absent uncertainty acknowledgment . 
Evaluation protocols and benchmarks such as FACTSCORE, FACTOR, and dialogue-level test sets distinguish supported versus unsupported claims and increasingly treat completeness as part of reliability, rather than focusing only on veracity \cite{kazlaris2025illusion,badawi2026assessing}. %By contrast, omission is less benchmarked, emphasizing the need for clinically grounded evaluation frameworks that jointly assess misinformation and missing safety-critical guidance\cite{badawi2026assessing}. % Mitigation approaches frequently involve prompt design and retrieval-augmented generation (RAG), but residual hallucination and omission remain when retrieved evidence is limited or incomplete. Overall, prior work positions hallucinations and omissions as core reliability risks for LLMs in high-stakes domains, where unsupported claims or missing guidance can induce harm \cite{alansari2025large}.

\textbf{LLMs for Mental Health Support and Safety Evaluation. } LLMs have increasingly been explored for mental health applications, %including educational resources and conversational support. %, due to their capacity for natural language interaction. Early reviews describe 
bringing both potential benefits for accessible mental health education, symptom assessment, and support, and risks when operated outside clinical supervision \cite{lawrence2024opportunities, xu2025evaluation}. % Prior work distinguishes general-purpose LLMs (e.g., GPT-family models) from domain-specific fine-tuned models with therapeutic objectives, with fine-tuned approaches often showing closer alignment with clinical guidance when trained on curated datasets \cite{boit2025prompt}. 
% Empirical studies assess LLM responses to mental health queries against clinical reference standards or professional guidelines, with attention to safety-relevant behaviors such as risk acknowledgement, appropriate empathy, escalation recommendations, and continued engagement. 
An empirical assessment of LLMs in mental health by Shah and colleagues indicated that there is substantial variability across models. \cite{shah2025evaluating}.  They further showed that some systems do not reliably provide actionable support or safe crisis responses . 
Systematic reviews from various disciplines, spanning psychiatry, psychology, psychotherapy, and clinical workflows, similarly 
concluded that, despite demonstrated technical feasibility, robust safety evaluation and clinical validation remain limited \cite{voultsiou2026systematic,badawi2026assessing}, highlighting ethical concerns and the need for interdisciplinary collaboration to reduce harm \cite{bucher2025s}. %This literature emphasizes involving clinical professionals in design and oversight, utilizing domain-specific evaluation metrics that prioritize relational sensitivity and risk escalation, and applying failure-aware evaluation frameworks that explicitly test crisis scenarios, contraindication guidance, and omission of essential safety information \cite{badawi2026assessing}. 
% Related reviews on AI in mental health further highlight ethical concerns including privacy, equity, informed consent, and the need for interdisciplinary collaboration to reduce harm \cite{bucher2025s}. Overall, prior work frames mental health LLM deployment as a clinical safety problem requiring dedicated benchmarks, human-centered evaluation protocols, and context-aware design \cite{lawrence2024opportunities}.

\textbf{Prompt Framing Effects on LLM Outputs. } % Prompt formulation and input framing are key determinants of LLM behavior. %, shaping response quality and the likelihood of failure modes such as hallucinations or unsafe omissions. 
Prior work shows that LLMs can be highly prompt-sensitive, where small changes in wording, structure, sentiment, or context encoding yield large differences in factuality, coherence, and bias, which can distort high-stakes decision tasks in ways unrelated to underlying evidence \cite{hwang2026wording}. 
%Studies of framing bias, analogous to psychological framing, find that wording can systematically shift model judgments even when semantic content is held constant, which can distort high-stakes decision tasks in ways unrelated to underlying evidence \cite{hwang2026wording}. 
Work on cultural value framing further demonstrates that prompt language and embedded cultural context can shift model responses and alignment with user values \cite{bulte2025llms}. 
% Prompt sensitivity also appears in formatting and structural variation (e.g., list versus paragraph), positional and order effects in few-shot prompts, and sentiment cues embedded in the input. These factors can materially affect model behavior, especially in domain-specific tasks where subtle signals influence generation \cite{gandhi2025prompt}. 
In addition, arbitrary prompt designs can induce statistical artifacts in LLM outputs, which threaten the internal validity of evaluations and motivate systematic prompt design protocols rather than ad hoc formulations \cite{brucks2025prompt}. 
% In healthcare and mental health contexts, prompt engineering research emphasizes bias-aware and safety-oriented framing. 
In mental health, variations in user background descriptions and contextual elements %, such as tone and emotional state, 
have been shown to affect empathy, risk recognition, and guidance on escalation, suggesting that neglecting framing elements can increase hallucinations and omissions in generated advice \cite{boit2025prompt}. % Overall, this literature indicates that prompt framing is a core determinant of LLM outputs with direct implications for reliability, safety, and ethical deployment in high-stakes settings.

\section*{Methods}

\paragraph{Evaluation corpus construction.}
We constructed an evaluation corpus of mental health inquiries using the $UTCO$ framework, which supports systematic and reproducible assessment of consumer-facing LLM behavior by varying clinically meaningful prompt elements while holding others constant. The $UTCO$ framework represents each inquiry by user background ($U$), clinical topic ($T$), situational context ($C$), and affective tone ($O$). This design enables controlled comparisons by varying one element while holding the others constant, and it broadens coverage of help-seeking language, including longer narratives and high-distress framing that are clinically important to evaluate\cite{marshall2024understanding}.

User background ($U$) comprised nine facet categories: user role, caregiver relationship, age group, life stage, gender identity, ethnicity or race, location or region, education level, and employment status. Clinical topic ($T$) was defined using a 10-domain taxonomy spanning depression- and anxiety-related concerns (e.g., Treatment Approaches, Crisis and Suicidality); full domain definitions and prevalence are reported in Table~\ref{tab:prevalence}. Situational context ($C$) was drawn from open-source peer support forums (Reddit and HealingWell) and curated scenario lists developed by the study team. Affective tone ($O$) was represented using 12 labels and each inquiry included zero to two tone labels: anxious, hopeless, frustrated, ashamed, guilty, angry, numb, resigned, urgent/crisis, confused, determined, and grateful\cite{kuvsen2017identifying}. In $UTCO$, $U$ captures user-background descriptors (roles and sociodemographics), whereas $C$ captures the situational narrative and circumstances surrounding the question.

Prompt generation followed a two-stage process. First, we sampled values for each $UTCO$ element from predefined discrete distributions, including a variable-length set of user background facets (zero to three per inquiry) and zero to two tone labels. We limited the number of background facets and tone labels to reflect how consumers typically present a small number of salient identity details and affect cues in a single help-seeking message, while still enabling controlled variation across elements. Second, each validated $UTCO$ combination was rendered into a single first-person inquiry using GPT-4o, constrained to one inquiry of at most 300 words and restricted to the sampled $UTCO$ fields. To reduce clinically or socially implausible combinations while retaining diversity, we applied an automated realism filter using GPT-4o (e.g., excluding a user younger than 18 years assigned a life stage of ``retired''), followed by expert review before downstream response generation and annotation. Experts screened for internal consistency across elements (e.g., age, life stage, and roles), clinically plausible topic-context pairings, and phrasing that remained consistent with a single first-person inquiry. 

Each inquiry was submitted to the target LLM (Llama 3.3, 70B) to generate a response. Three independent annotators labeled each inquiry-response pair for hallucination and omission, with disagreements resolved through adjudication by a senior medical informatics expert team. Outcomes were binary indicators. We defined a \emph{hallucination} as the inclusion of clinically incorrect statements or fabricated medical resources not supported by the inquiry\cite{jin2025exploring}, and an \emph{omission} as failure to provide clinically substantive or safety-critical content that would reasonably be expected given the inquiry, \emph{including essential safety guidance even when not explicitly requested (e.g., crisis resources in self-harm related inquiries)}\cite{salleh2023errors}.

\paragraph{Feature representation.}
We transformed each $UTCO$ inquiry into a structured feature set for modeling hallucination and omission outcomes. User background facets (e.g., caregiver, age group, education level) and tone labels (e.g., anxious, hopeless) were encoded as binary indicators. Clinical topic was represented using indicator variables for both the topic domain and its corresponding subtopic in the hierarchical taxonomy. Because situational context was free text, we derived context-level linguistic features that quantify how the inquiry is expressed, including whether the context source was naturalistic, context word count, readability, sentiment, passive-voice and subordinate-clause counts, medical-term density, risk-term density, an uncertainty-term score, and a pronoun ambiguity score. For example, one inquiry could be encoded with indicators such as $U$=caregiver, $T$=Crisis and Suicidality, $O$=hopeless, together with continuous context features such as context word count and readability. Indicators with near-zero prevalence were retained for regularized models and excluded from unregularized models to reduce unstable estimates.

\paragraph{RQ1: Feature association and global risk profiling.}
To identify which UTCO-derived features were most associated with failure risk, we fit separate predictive models for hallucination and omission. For each outcome, we trained a gradient-boosted tree classifier using the same preprocessed feature set, then computed SHAP values to summarize global feature contributions across inquiries. We interpreted SHAP values as global contributions to predicted risk rather than as statistical significance.

\paragraph{RQ2: Sensitivity analysis via propensity score matching.}
To formally test whether failures were more consistent with differences in \emph{who the user is} (user background) versus \emph{how the inquiry is expressed} (linguistic characteristics), we performed a four-round leave-one-out propensity score matching analysis. In each round, %and outcome, 
prompts with the failure were treated cases and prompts without the failure were controls. We estimated the probability of failure using a logistic regression propensity model conditioned on three $UTCO$ elements being balanced in that round, with covariates standardized before fitting. We then performed 1:1 nearest-neighbor matching on propensity score without replacement and retained pairs only when the absolute propensity score difference was below a prespecified threshold.

Within matched pairs, we compared the held-out $UTCO$ element between treated and control prompts. For binary indicators (e.g., user background facets or tone labels), we applied McNemar's test and reported paired prevalence differences (treated minus control). For continuous context features, we used paired t-tests and reported paired mean differences. Under this design, differences observed in the held-out element after balancing the other three elements provided evidence consistent with the sensitivity of failure risk to that element.

\paragraph{RQ3: Similarity-matched mechanism analysis.}
To identify residual triggers not captured by the $UTCO$ elements, we paired each failure inquiry with highly similar non-failure inquiries using the same $UTCO$-derived feature set. We standardized the feature matrix using z-score scaling and, for each hallucination or omission case, retrieved up to $K=10$ nearest non-failure controls using cosine distance. We retained neighbors only when cosine distance was at most 0.15.

We used GPT-4o as a structured judge to scale within-pair comparisons under a fixed rubric, and we validated this judge against human ratings to assess reliability. For each retained pair, the judge evaluated differences using a predefined six-dimension taxonomy, \emph{ambiguity, contradiction, missing constraints, temporal confusion, multi-intent, and emotional load}, and assigned ordinal severity scores on a 0 to 3 scale. The judge also extracted short evidence spans and proposed a minimal counterfactual rewrite that preserved intent while reducing the identified trigger. To validate the LLM-as-a-judge scoring, we manually reviewed a random sample of 100 similarity-matched pairs. Two annotators (CN, SQ) independently rated each pair on the same six mechanism dimensions (ambiguity, contradiction, missing constraints, temporal confusion, multi-intent, and emotional load) using the 0 to 3 ordinal rubric. We summarized concordance using within-one-point agreement, defined as an absolute difference of at most one severity level.

\section*{Results}
\paragraph{Cohort characteristics.}
The final analytic cohort included $n=2,075$ mental health inquiries generated using the $UTCO$ framework. Across all model responses, hallucinations occurred in 6.50\% (n=134) and omissions occurred in 13.20\% (n=273). User background facets were encoded as binary indicators and were optionally included on a per-inquiry basis (zero to three facets), so the percentages below reflect the share of inquiries that explicitly contained each facet rather than a complete demographic distribution. For example, inquiries that specified gender identity included women (1.90\%), men (1.70\%), trans women (1.60\%), and non-binary individuals (1.60\%). Inquiries that specified age ranged from 18 to 24 years (2.00\%) to 65+ years (1.50\%). Caregiver (2.70\%) and parent (2.50\%) roles were also represented.

Clinical topics spanned 10 domains, with domain-specific failure prevalence summarized in Table~\ref{tab:prevalence}. The most common domain was \emph{Clinical Depression Experience and Management} (n=348). Hallucinations were most frequent for \emph{Medication-Specific Issues} (10.90\%), whereas omissions were most frequent for \emph{Crisis and Suicidality} (36.20\%).

Situational context was summarized using nine context-level linguistic features derived from the context text, including naturalistic-source status, readability, sentiment, passive voice, subordinate clauses, medical-term density, risk-term density, uncertainty, and pronoun ambiguity. Naturalistic context sources accounted for 29.10\% of prompts. Context readability (Flesch-Kincaid) had a mean of 9.70$\pm$5.90 (range 0.00 to 32.00), and context sentiment ranged from -0.90 to +0.80 (mean 0.01$\pm$0.18). Passive constructions were uncommon and appeared as a binary signal (4.20\% with at least one passive nominal subject).

Tone was represented using 12 affect indicators. High-distress tones were common, including ``Hopeless'' (13.50\%), ``Anxious'' (13.00\%), and ``Confused'' (11.10\%), while 19.60\% of prompts did not include a tone indicator.

\begin{table}[ht]
\centering
\caption{Prevalence of hallucinations and omissions by clinical topic domain ($N=2{,}075$).}
\label{tab:prevalence}
\setlength{\tabcolsep}{6pt}
\renewcommand{\arraystretch}{1.15}
\begin{small}
\begin{tabular}{|l|c|c|c|}
\hline
\textbf{Clinical topic domain} & \textbf{Total $N$} & \textbf{Hallucination rate} & \textbf{Omission rate} \\
\hline
Medication-Specific Issues & 174 & 10.9\% (19) & 13.2\% (23) \\
\hline
Professional Care Navigation & 155 & 9.7\% (15) & 10.3\% (16) \\
\hline
Anxiety Experience and Management & 300 & 8.3\% (25) & 14.0\% (42) \\
\hline
Clinical Depression Experience and Management & 348 & 7.5\% (26) & 18.7\% (65) \\
\hline
Social \& Relationship Dynamics & 179 & 7.3\% (13) & 7.3\% (13) \\
\hline
Treatment Approaches & 272 & 6.6\% (18) & 9.2\% (25) \\
\hline
Crisis and Suicidality & 163 & 4.3\% (7) & 36.2\% (59) \\
\hline
Specific Comorbidities & 189 & 2.6\% (5) & 3.7\% (7) \\
\hline
Identity and Self-Perception & 184 & 2.2\% (4) & 8.7\% (16) \\
\hline
Administrative \& Legal Concerns & 98 & 0.0\% (0) & 6.1\% (6) \\
\hline
\end{tabular}
\end{small}
\end{table}

\paragraph{RQ1: Feature prevalence and predictors of failure.}

\begin{figure}[ht]
    \centering
    \includegraphics[width=1\textwidth]{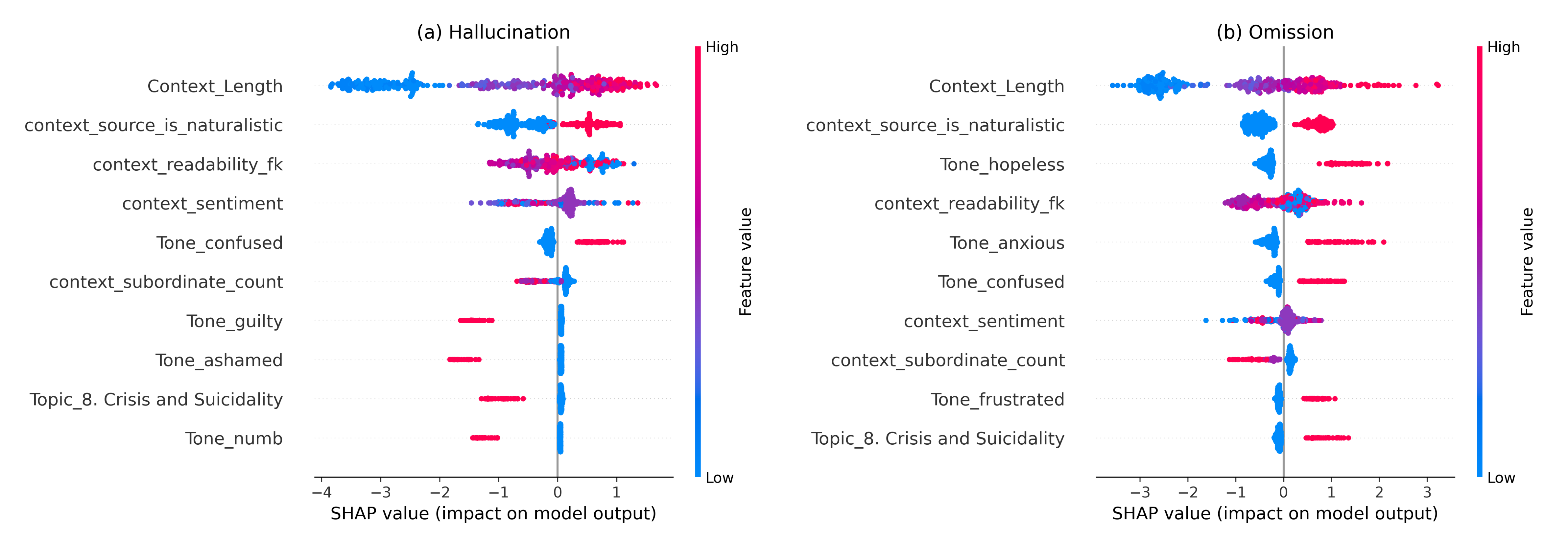}
    \caption{SHAP summary plots for the gradient-boosted tree models predicting hallucination (a) and omission (b). Features are ordered by overall contribution magnitude across inquiries.}
    \label{fig:shap}
\end{figure}

\paragraph{Hallucination.} In the hallucination model, SHAP rankings (Figure~\ref{fig:shap}a) were dominated by context length (prompt word count) and naturalistic-source status, indicating higher predicted risk for longer prompts and prompts drawn from naturalistic contexts. Readability and sentiment showed smaller contributions, suggesting that both the amount of contextual information and its linguistic framing can shift model reliability. Among tone indicators, \emph{confused} contributed toward higher predicted risk, whereas \emph{guilty}, \emph{ashamed}, and \emph{numb} tended to contribute toward lower predicted risk. Topic indicators, including \emph{crisis and suicidality}, also appeared among top contributors.

\paragraph{Omission.}
In the omission model, SHAP rankings (Figure~\ref{fig:shap}b) similarly prioritized prompt word count and naturalistic-source status, consistent with higher predicted omission risk in longer, narrative prompts. High-distress tone indicators were among the strongest contributors, particularly \emph{hopeless}, \emph{anxious}, and \emph{confused}. Additional context features, including readability, sentiment, and subordinate-clause count, contributed at smaller magnitude, consistent with an added effect of syntactic complexity and affective framing. Topic indicators, including \emph{crisis and suicidality}, also contributed to predicted omission risk.

\paragraph{RQ2: Sensitivity analysis via propensity score matching.}
Across leave-one-out matching rounds, we found limited evidence of residual differences attributable to user background, topic, or tone after balancing the remaining $UTCO$ elements. Specifically, when user background was held out (matching on topic, context, and tone), paired comparisons of user background indicators did not differ between failure cases and matched controls for hallucination or omission (all $p>0.05$). Similarly, when topic or tone was held out, we did not observe statistically significant differences in topic or tone indicators between failures and matched controls (all $p>0.05$). Under the matched design, these results did not support the sensitivity of failure risk to user demographic profiles, topic selection, or tone labels.

By contrast, when context was held out (matching on user background, topic, and tone), context-level linguistic characteristics differed consistently between failures and matched controls. For hallucinations ($n=76$ matched pairs), failure cases had higher readability grade level (12.70 vs.\ 10.10, $p<0.001$) and longer prompt word count (21.90 vs.\ 17.10, $p<0.001$). For omissions (n=135 matched pairs), failure cases similarly showed higher readability grade level (12.40 vs.\ 9.70, $p<0.001$) and longer prompt word count (21.60 vs.\ 12.90, $p<0.001$). Omission cases also had higher pronoun ambiguity (0.40 vs.\ 0.18, $p=0.002$), more subordinate clauses (0.53 vs.\ 0.29, $p=0.003$), higher uncertainty score (0.0040 vs.\ 0.0007, $p=0.009$), and higher medical-term density (1.20\% vs.\ 0.60\%, $p=0.018$). Overall, these matched-pair differences indicate that failures were most sensitive to how contextual information was expressed, with omissions showing additional sensitivity to ambiguity and clinical specificity.

\paragraph{RQ3: Similarity-matched mechanism analysis.}

\paragraph{Judge validation.}
Across 100 similarity-matched pairs and six ordinal dimensions (600 ratings), within-one-point agreement between the two annotators (CN, SQ) was 78.33\% (470/600). Agreement between the GPT-4o judge and the pooled human reference score (rounded mean of the two annotators) was 92.83\% within one point (557/600).

\paragraph{Match yield and similarity.}
Since each failure case can be paired with up to $K=10$ non-failure controls, we summarize the results at the case level. We imposed a cosine distance threshold of $d \le 0.15$ and retained only controls within this threshold, such that the realized number of qualifying controls varied by case. Under this threshold, 61/134 (45.50\%) hallucination cases and 66/273 (24.20\%) omission cases had at least one qualifying non-failure neighbor, indicating that many failures occurred in regions of the $UTCO$ feature space where no sufficiently similar non-failure inquiries were available. Among retained cases, similarity remained high: the median case-to-control cosine distance (averaged over retained controls per case) was 0.079 (IQR 0.050 to 0.103) for hallucinations and 0.087 (IQR 0.039 to 0.122) for omissions.

% \begin{figure}[t]
% \centering
% \includegraphics[width=0.9\textwidth]{Figures/rq3_signature_dotplot.png}
% \caption{\textbf{Failure-mode linguistic signatures.} Mean (points) and standard deviation (error bars) of case-level LLM-judge severity scores (0 to 3) for hallucination vs.\ omission outcomes under $UTCO$ similarity matching (up to $K=10$ non-failure controls per case; $d \le 0.15$).}
% \label{fig:rq3_signature}
% \end{figure}

\begin{figure}[h!]
\centering
\includegraphics[width=\textwidth]{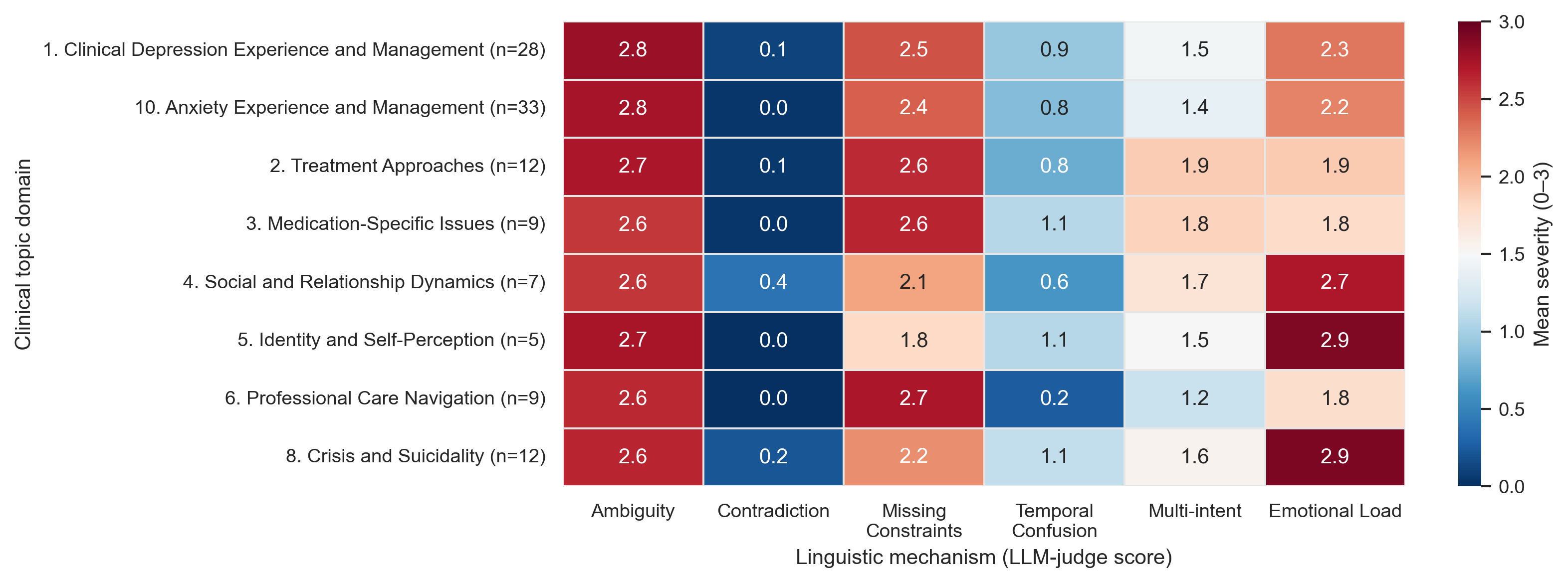}
\caption{\textbf{Topic-stratified linguistic trigger profiles (descriptive).} Heatmap of case-level mean LLM-judge severity scores (0 to 3) across topic domains. Topic-level summaries are descriptive and may be unstable in sparsely represented domains.}
\label{fig:rq3_heatmap}
\end{figure}

\paragraph{Trigger profiles differed by failure mode.}
The LLM judge assigned ordinal severity scores (0 to 3) across six mechanism dimensions (Table~\ref{tab:rq3_taxonomy_scores}). Across both failure modes, \emph{ambiguity} showed the highest mean severity (hallucination: 2.77$\pm$0.37; omission: 2.67$\pm$0.45), followed by \emph{missing clinical constraints} (hallucination: 2.65$\pm$0.44; omission: 2.26$\pm$0.58). The profiles differed most clearly on distress-related features. \emph{Emotional load} was higher for omissions than hallucinations (2.46$\pm$0.73 vs.\ 2.00$\pm$0.76), indicating greater sensitivity to distress cues even under high-similarity matching. In contrast, hallucinations had higher \emph{missing clinical constraints} scores, consistent with failures arising when key clinical details were underspecified. \emph{Logical contradiction} remained near zero for both modes (hallucination: 0.03$\pm$0.13; omission: 0.15$\pm$0.40).

\paragraph{Topic-stratified patterns.}
Figure~\ref{fig:rq3_heatmap} shows that mechanism scores differed across clinical topics. Across topics, \emph{ambiguity} and \emph{missing clinical constraints} were consistently high (typically $\ge 2.10$), while omission cases often showed higher \emph{emotional load} (approaching 2.70 to 2.90) and \emph{multi-intent} varied more modestly (approximately 1.20 to 1.90).

\paragraph{Representative matched examples (narrative).}
Representative similarity-matched pairs illustrated recurring mechanisms under high structural similarity. For ambiguity and missing constraints, failures often involved short, underspecified questions (e.g., ``How long does depression last?'') relative to a successful neighbor with clearer scope, and judge-proposed rewrites added a minimal constraint (e.g., specifying ``a depressive episode'' and the target population). For emotional load, omission failures were frequently distinguished by high-distress framing (e.g., references to persistent panic and fear of ``never feeling normal''), and rewrites reduced urgency cues while preserving the request for coping strategies. For multi-intent, failures often combined multiple goals in one inquiry (e.g., symptom description plus logistical questions about care platforms), and rewrites narrowed the prompt to a single actionable intent (e.g., locating affordable counseling given specific constraints).

\begin{table}[ht]
\centering
\caption{\textbf{Linguistic taxonomy severity by failure mode (case-level).} Scores (0 to 3) adjudicated by a blinded judge across up to $K=10$ neighbors. Values are reported as mean$\pm$SD.}
\label{tab:rq3_taxonomy_scores}
\setlength{\tabcolsep}{7pt}
\renewcommand{\arraystretch}{1.15}
\begin{small}
\begin{tabular}{|l|c|c|}
\hline
\textbf{Linguistic Dimension (0--3)} & \textbf{Hallucination (n=61)} & \textbf{Omission (n=66)} \\
\hline
Ambiguity / Underspecification & $2.77 \pm 0.37$ & $2.67 \pm 0.45$ \\
\hline
Missing Clinical Constraints & $2.65 \pm 0.44$ & $2.26 \pm 0.58$ \\
\hline
Emotional Load / Crisis Cues & $2.00 \pm 0.76$ & $2.46 \pm 0.73$ \\
\hline
Multi-intent / Scope Creep & $1.57 \pm 0.75$ & $1.46 \pm 0.80$ \\
\hline
Temporal Confusion & $0.76 \pm 0.56$ & $0.89 \pm 0.58$ \\
\hline
Logical Contradiction & $0.03 \pm 0.13$ & $0.15 \pm 0.40$ \\
\hline
\end{tabular}
\end{small}
\end{table}

\section*{Discussion and Conclusions}

% \subsection*{Key findings and contributions}
% In a cohort of 2,075 synthetic mental health inquiries, hallucinations occurred in 6.5\% (n=134) and omissions occurred in 13.2\% (n=273). Failure prevalence varied substantially across clinical topic domains, with hallucinations highest in Medication-Specific Issues (10.9\%) and omissions highest in Crisis and Suicidality (36.2\%). These findings indicate that failure risk is not uniformly distributed, and that clinically sensitive topics concentrate omission risk.

% This study proposes an evaluation design that treats prompt construction as an analyzable safety dimension. By combining (1) multivariate association modeling, (2) leave-one-out propensity score matching across $UTCO$ elements, and (3) similarity-based pairing and mechanism scoring, these analyses collectively reveal areas where failures are concentrated, which prompt features remain associated with failures under tightly controlled conditions, and residual linguistic patterns that emerge in nearly matched comparisons. This structure supports practical stress testing of patient-facing large language models for mental health across user background, topic, context, and tone.

\textbf{Omissions warrant equal attention as hallucinations.} In this study, omissions were more common than hallucinations (13.2\% vs.\ 6.5\%), and the difference between these two failure rates was larger in clinically sensitive content. This matters because omission is often a \emph{silent} failure mode. Unlike hallucinations, which may be detectable as overtly incorrect statements, omissions can produce responses that appear coherent and empathic while still failing to deliver clinically necessary elements, such as risk escalation guidance, safety checks, or concrete next steps. In mental health settings, this makes omissions harder for users to recognize and easier to act on, particularly under distress. The practical implication is that omission should be treated as a primary safety outcome in mental health LLM evaluation, not as a secondary complement to hallucination. This emphasis is consistent with the ethical principles of beneficence and non-maleficence, which prioritize providing benefit while reducing avoidable harm. Omissions can also undermine autonomy by withholding information needed for informed decision-making, particularly in safety-critical situations\cite{gillon1994medical,beauchamp2019principles}.

Across analyses, we found that omission risk was most consistently associated with prompt characteristics that resemble real-world help-seeking messages. In regression analyses, longer contextual narratives and \emph{naturalistic context sources} were associated with higher omission risk, and prompts expressing high emotional distress were also associated with omission. Element-specific matching analyses further showed that, when user background and topic were held constant, omission cases differed from control prompts in several features indicating greater cognitive burden (i.e., longer context, higher readability levels and higher rates of pronoun ambiguity). This pattern is consistent with the observation that omissions are concentrated in prompts related to crisis and suicidal ideation, where users often provide narrative context and an urgent emotional framework but may not provide explicit clinical constraint\cite{pichowicz2025performance}. These results, consistent with prior observations of omission patterns in crisis-related prompts, highlight a gap between simplified benchmarking prompts and the complex narratives that characterize real-world help-seeking.

These findings offer two key implications for the assessment and design of consumer health informatics. First, benchmarking with short, concise prompts may underestimate the risk of information omissions in mental health applications, as it fails to reflect the lengthy and ambiguous narratives commonly found when seeking help in the real world. Therefore, we recommend assessment protocols that incorporate stress testing, systematically altering context length, contextual sources, and emotional tone. Furthermore, such protocols should use clinically guided checklists to assess information omissions and focus on safety-related content rather than general utility. Second, mitigation strategies should prioritize preventing information omissions in high-risk scenarios, such as implementing structured safety information supplementation steps when crisis indicators appear and triggering clarification questions when key limitations are missing or pronouns are severely ambiguous. This shifts the focus from generating fluent, supportive text to reliably providing the minimum safety information required for consumer-facing mental health support.

\textbf{Context and tone shift failure risk. } Across our analyses, failure risk was most sensitive to \emph{how the inquiry was framed and narrated} rather than to the user-background element itself. Context characteristics and affective tone repeatedly emerged as stable risk signals, while user-background indicators did not show systematic differences after balancing other elements. In practical terms, this suggests that in mental health question answering the model is more likely to fail when the prompt contains complex information or strong affect, that is, when users describe symptoms and needs in variable, ambiguous, or emotionally distressed ways, as is common in consumer-facing systems.

 A plausible explanation for this observation is that longer, more naturalistic narratives increase the difficulty of the model's comprehension\cite{jiang2024longllmlingua}. These prompts often require the model to infer missing clinical constraints, resolve ambiguous references, and integrate multiple clues into coherent recommendations. When the tone indicates high patient anxiety, the prompts also carry a stronger emotional load and sense of urgency, further prompting the model to provide broad, empathetic responses but lacking specific guidance. In this scenario, hallucinations and omissions may stem from different failures: hallucinations might reflect the model's overconfidence in completing information under uncertainty\cite{simhi2025trust}, while omissions might reflect the model's failure to translate narrative and emotions into necessary safety-related content\cite{Jiao2025_NavigatingLLMEthics}.

These findings lead to two key insights. First, evaluation protocols should prioritize stress tests that can clearly alter context length, narrative style, and emotional tone, rather than relying solely on short, well-defined prompts. Second, mitigation strategies should focus on ``uncertainty management'' and ``safe completion'' in situations involving narrative ambiguity and emotional distress. Specifically, the system can trigger clarifying questions when key constraints are missing, detect ambiguous reference patterns, and enforce a set of basic safety measures when crisis signals appear, even if the prompt does not explicitly request them.

\textbf{Limitations and future work. } While the $UTCO$ inquiry design improved scalability and experimental control, it may not fully represent how people describe mental health concerns in real-world help-seeking, including in clinical encounters or peer support settings. Future research should compare the distribution of $UTCO$ inquiries with naturally occurring inquiries and assess whether $UTCO$ coverage aligns with real-world topic, context, and tone patterns. In addition, our assessment focused on a single target model configuration (Llama 3.3, 70B). Although we observed consistent risk patterns within this model, their generalizability to other LLM families and deployments requires direct replication. Mechanistic analysis was also constrained by the limited availability of highly similar non-failure neighbors under the cosine distance threshold, particularly for omissions, where only 66/273 cases (24.20\%) had at least one qualifying control. Future research should extend mechanism testing through \emph{systematic manipulation} of specific linguistic features (e.g., ambiguity, constraints, and multi-intent structure) and estimation of how targeted perturbations alter hallucination and omission risk. Finally, our realism filter and expert review emphasized internal consistency across $UTCO$ elements, which may have reduced the prevalence of logically contradictory inquiries and contributed to near-zero contradiction scores in the mechanism analysis. Future work should stress test contradictions by introducing controlled inconsistencies in otherwise realistic narratives.

% \section*{Conclusion}
% In this study, we evaluated Llama 3.3 on 2,075 structured mental health inquiries generated with the $UTCO$ framework and quantified two clinically relevant failure modes: hallucinations and omissions. Hallucinations occurred in 6.5\% of responses and omissions occurred in 13.2\%, with omission risk concentrated in clinically sensitive scenarios, including crisis and suicidal ideation prompts. By combining multiple complementary analytical methods, including relational modeling, specific element matching, and high-similarity comparison, we found that failures were most closely associated with contextual and tonal features, while user background indicators did not show systematic differences after balancing other prompt elements. These findings collectively suggest that fixed and explicit benchmarks alone are insufficient to adequately assess the safety of large language models for mental health applications. Therefore, it is necessary to develop evaluation frameworks that explicitly stress-test narrative context and emotional framing, and conduct mechanism-oriented analyses to identify subtle prompt features that trigger unsafe or incomplete responses.

\section*{Funding Acknowledgement}
This research was, in part, funded by the Advanced Research Projects Agency for Health (ARPA-H). The views and conclusions contained in this document are those of the authors and should not be interpreted as representing the official policies, either expressed or implied, of the United States Government.

\makeatletter
\renewcommand{\@biblabel}[1]{\hfill #1.}
\makeatother
\bibliographystyle{plain}
\bibliography{amia}

\end{document}